\documentclass[letterpaper, 10 pt, conference]{ieeeconf}  

\usepackage{graphicx}
\usepackage{xcolor}
\usepackage{textcomp}
\usepackage{amsmath}
\usepackage{booktabs}
\usepackage{gensymb}
\usepackage{comment}
\usepackage[ruled,vlined,linesnumbered]{algorithm2e}
\usepackage{amsmath}
\usepackage[normalem]{ulem}

\SetKwComment{Comment}{/* }{ */}
\SetKwFor{ForEach}{foreach}{do}{end foreach}
\newcommand{\var}[1]{\texttt{#1}}

\usepackage[hyphens]{url}
\usepackage[colorlinks=false,pdfborder={0 0 0}]{hyperref}

\IEEEoverridecommandlockouts                              

\title{\LARGE \bf
HERB: Human-augmented Efficient RL for Bin-packing}

\author{Gojko Perovic$^{1}$  Nuno Ferreira Duarte$^{2}$  Atabak Dehban$^{2}$  Gonçalo Teixeira$^{2}$
 \\
 Egidio Falotico$^{1}$ José Santos-Victor$^{2}$
\thanks{Corresponding author - GP: {\tt\small{g.perovic@santannapisa.it}}}
\thanks{$^{1}$BioRobotics Institute, Scuola Superiore Sant’Anna, Pisa, Italy, and with the Department of Excellence in Robotics and AI, Scuola Superiore Sant’Anna, Pisa, Italy. \{\tt\small{{g.perovic, e.falotico\}@santannapisa.it}}}
\thanks{$^{2}$Institute for Systems and Robotics, Instituto Superior Técnico, Universidade de Lisboa, 1049-001 Lisbon, Portugal. 
\{\tt\small{{nferreiraduarte, adehban,jasv\}@isr.tecnico.ulisboa.pt}}}
\thanks{This work was supported by the LARSyS FCT funding -10.54499/UIDB/50009/2020, the Lisbon ELLIS unit, the Center for Responsible AI, and MPR-2023-12- SACCCT- Project 14935 AI.PackBot.}
}

\begin{document}

\maketitle
\thispagestyle{empty}
\pagestyle{empty}

\begin{abstract}

Packing objects efficiently is a fundamental problem in logistics, warehouse automation, and robotics. 
When dealing with highly diverse 3D objects (household or grocery items), closed-form solutions are infeasible, and heuristic or model-free Reinforcement Learning~(RL) methods tend to focus solely on geometric optimization, relying on exhaustive searches of the discretized solution space. This leads to long training times (for pure RL) and high latency (heuristics), limited transferability to robotic scenarios, and ultimately ignores object characteristics (fragility, deformability) and human preferences.

We propose HERB, a human-augmented RL framework for packing irregular objects, the first to explore the potential of learning from human demonstrations to solve this complex task.
It leverages human demonstrations of packing strategies, which inherently exhibit latent factors such as space optimization, stability, and object properties that are difficult to model explicitly. The human-expert data is combined with RL exploration to provide the placement of each object inside the container. 
Experimental results show that our method outperforms heuristic, purely RL-based, and imitation learning approaches in packing efficiency and latency. Qualitative results highlight that our packing strategy produces more stable, human-like arrangements, which we expect to be more appropriate and widely accepted. 
Finally, we demonstrate the real-world feasibility of our method on a robotic system.
%
\end{abstract}

\section{INTRODUCTION}

Irregular object packing is a fundamental skill for robotic applications across both industrial and household domains and is often formulated as a bin-packing problem addressed with optimization techniques. 
However, even the two-dimensional discrete version of the bin-packing problem, restricted to regular objects, has been proven to be NP-hard~\cite{clautiaux_new_2007}.
Despite the inherent difficulty of this task, numerous solutions and variations have been explored in the literature~\cite{pantoja-benavides_comprehensive_2024, leao_irregular_2020, coffman_approximation_1984}. When dealing with objects of arbitrary irregular shapes, the demands on the robotic system’s perception, planning, and manipulation capabilities increase substantially~\cite{pantoja-benavides_comprehensive_2024, leao_irregular_2020}, and both a challenge and an opportunity for developing more robust and adaptable solutions are posed.

As closed-form solutions are not available, recent approaches for irregular object packing typically rely on either heuristic rules or RL~\cite{goyal_packit_2020, huang_planning_2023, zhao_learning_2023}. 
However, these methods suffer from two major limitations. 
First, they depend heavily on rule or reward design, often focused on geometric optimization alone, which can compromise performance in real-world robotic scenarios.
Second, by discretizing the solution space and relying on exhaustive search, these approaches both restrict the range of possible solutions and reduce efficiency, because of slow inference for heuristic methods and prolonged training times for RL-based methods.

Humans, by contrast, can efficiently pack multiple objects into a box while simultaneously accounting for a variety of constraints, such as object category or whether one object can be placed atop another. 
Observing human behavior not only offers insights for solving the geometric aspects of the problem but also provides access to latent object properties (such as fragility, deformability, or compatibility) that are otherwise unavailable to a robotic system. 
Leveraging this information could help overcome the limitations of RL or heuristic-based approaches, enabling more efficient and successful completion of the packing task.
\begin{figure}[t]
    \centering
    \includegraphics[width=\linewidth]{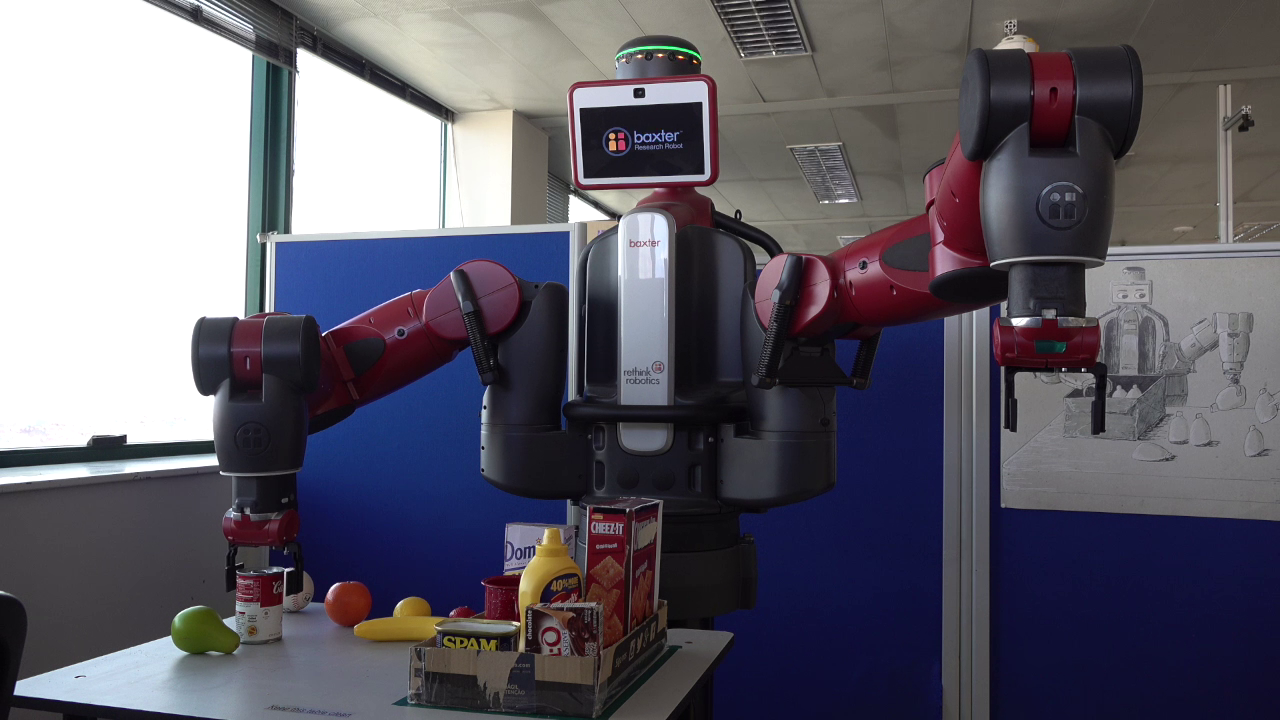}
    \caption{Baxter\textregistered\ robot with the box and objects from the BoxED dataset. The heightmap images capturing the state of the box are taken with a depth camera~(out of crop).}
    \label{fig:setup}
\end{figure}

Learning from humans has driven many recent advances in robotic manipulation~\cite{zare_survey_2024}. 
In a similar vein, we leverage human demonstrations to frame and tackle the problem of irregular object packing as a robotic challenge in continuous space. 
In its most common form, learning from human demonstration, also known as Imitation Learning (IL), is approached as a supervised learning problem commonly referred to as Behavioral Cloning (BC). 
However, certain aspects of irregular object packing, such as placement prediction, are difficult to model via IL due to the highly diverse solution space and stringent accuracy requirements. 
Since irregular object packing has not been extensively explored with such methods, we conduct a study using two representative IL models to highlight these challenges and introduce a strategy to overcome them.

We address the problem of irregular 3D object bin packing using a hybrid learning approach that combines IL with RL. 
In our framework, the discrete components of the task, which involve a relatively small search space (i.e., sequence planning), are learned from human demonstrations, while the continuous, high-dimensional actions (including the precise placement location and orientation within the box) are learned via RL. 
Our main contributions are threefold:

\begin{itemize} \setlength\itemsep{0em}
    \item We introduce HERB (Human-augmented Reinforcement learning-based Bin-packing), which integrates a human-like sequence prediction algorithm with RL to tackle irregular object packing. HERB outperforms state-of-the-art heuristic, purely RL-based, and imitation learning approaches in common household object packing tasks.
    \item We develop a lightweight simulation environment and a corresponding dataset of human packing demonstrations, enabling efficient training and evaluation of different methods. Both the environment and dataset will be made publicly available in the final version of the paper.
    \item To validate HERB and qualitatively assess its packing performance, we implement a full robotic pipeline using a Baxter\textregistered robot as the manipulation platform (Fig.~\ref{fig:setup}).
\end{itemize}

\section{RELATED WORK}
\label{sec:rw}

Object packing has been a long-standing challenge due to its high practical significance and inherent complexities~\cite{pantoja-benavides_comprehensive_2024,  leao_irregular_2020, coffman_approximation_1984}.
In the most common definition, the planning segment of the problem is referred to as the Bin Packing Problem~\cite{coffman_approximation_1984}, where sets of cuboid objects must be placed in discrete bins of a larger container.
As we aim to contextualize object packing as a skill for a robotic manipulator, we consider irregular 3D objects and a continuous placement space.

Under these considerations, closed-form solutions for irregular object packing are not available.
Thus, in 3D, irregular object packing planning methods can largely be divided into \textit{Heuristic} or \textit{RL-based} methods:

\textbf{Heuristic solutions} aim to solve the packing problem by employing geometrical estimates of the optimal object placement. While performing well, these methods are computationally expensive in the inference phase~\cite{goyal_packit_2020, pan_sdf-pack_2023}.
Furthermore, by using heuristics, the solution space can become overly constrained, which may lead to difficulties when applying it to real-world scenarios.
For example, the exact 3D models of the objects may be missing, or the perception framework may not be precise enough to observe the voxel occupancy of the box.
More critically, in what concerns practical scenarios, these methods tend to strictly focus on solving the combinatorial problem and space optimization to the detriment of the physical feasibility of the solution.

\textbf{RL-based solutions} aim to use a data-driven approach to solve the planning problem~\cite{huang_planning_2023, zhao_learning_2023, hu_solving_2017}.
In the case of regular cuboid objects and discrete bins, straightforward RL solutions tend to work well~\cite{hu_solving_2017}.
When irregular objects are considered, the algorithm design becomes significantly more complex, leading to larger models~\cite{huang_planning_2023, zhao_learning_2023}.
Thus, RL-based methods rely on either discretizing the action space by binning~\cite{huang_planning_2023} or employing a geometric-based heuristic~\cite{zhao_learning_2023} and then running an exhaustive search to estimate the state-action value $Q$ of candidate solutions.
While such discretizations simplify the RL approach, they constrain the method's potential generalization capabilities and, in many cases, exhaustive search leads to longer training times~\cite{huang_planning_2023, zhao_learning_2023}.


\textbf{Learning from humans} has emerged as one of the leading paradigms when considering data-driven approaches for acquiring complex robotic skills~\cite{zare_survey_2024}.
While human-inspired approaches are at the forefront of enabling complex robotic policies, they have not been exploited for 3D irregular object packing so far.
Santos et al.~\cite{santos_andrejfsantos4boxed_2025} presented the \textit{Box packing with Everyday items Dataset (BoxED)}.  
In this study, the dataset is used to enable human-like packing policies, as well as to assess the performance of different approaches and compare them with human packing, both in terms of packing utility and qualitative characteristics. 
To facilitate the training and evaluation of different methods, a \textit{PyBullet} environment based on the dataset collection experiment~\cite{santos_andrejfsantos4boxed_2025} is implemented. 
A method for human-like packing sequence generation has also been described in the literature~\cite{santos_learning_2024}. 
In this study, a variant of that algorithm is adapted to alleviate the training and implementation complexity of the RL algorithm and to generate object packs with human-like qualities. 
Accordingly, the packing problem is decoupled into two sub-tasks: Sequence Planning (ordering the objects to pack) and Placement Prediction (how to place them).

\section{LEARNING PACKING FROM HUMAN DEMONSTRATIONS}
\label{sec:approach}

\subsection{Human-like Sequence Planning}

Given a set of available objects $O = \{obj_A, obj_B, ...\}$, and the dataset of human packing sequences~\cite{santos_andrejfsantos4boxed_2025}, a model of human-like transitions can be developed.
More specifically, a static Markov Chain based on pairwise transitions $obj_A \rightarrow obj_B$ is reconstructed~\cite{santos_learning_2024}.
By applying a modified beam-search algorithm, the obtained transition matrix can be sampled to generate human-like sequences.
Furthermore, experiments have shown that constraining the modified beam search to sample the next 3 objects (referred to as Beam-3) produces the most human-compatible sequences \cite{santos_learning_2024}.

Thus, given a list of objects, we generate a human-like sequence using the Beam-3 method.
The ordered sequence should not only be beneficial in terms of size and geometry compatibility, but importantly, implicit information in human sequence preferences can be exploited.
The proposed sequence planning algorithm is defined in Algorithm \ref{al:beam}.
\begin{algorithm}[t]
\DontPrintSemicolon
\caption{Beam-3 Algorithm}
\label{al:beam}
\small
\var{beam\_width} $\gets 5$\;
\var{max\_length} $\gets 3$\;
\var{all\_transitions} $\gets$ \{transition probabilities from the human packing dataset\}\; 
\var{unsorted\_objects} $\gets$ \{all available objects\}\;
\var{sorted\_sequence} $\gets \emptyset$\;

\While{\var{unsorted\_objects} $\neq \emptyset$}{
    \var{beams} $\gets$ \{initial beam with $\langle \text{start} \rangle$ state\}\;
    
    \Comment{\footnotesize Beam search to find next object}
    \While{true}{
        \var{has\_expansion} $\gets$ false\;
        
        \ForEach{\var{beam} in \var{beams}}{
            \If{length(\var{beam}) $\geq$ \var{max\_length}}{
                \textbf{continue};}
            \Else{
                \var{valid\_transitions} $\gets$ Gen(\var{beam}, \var{all\_transitions})\; \Comment{\footnotesize Get available transitions}
                \If{\var{valid\_transitions} $\neq \emptyset$}{
                    \var{beams} $\gets$ \var{beams} $\cup$ Branch(\var{beam}, \var{valid\_transitions})\;
                    \var{has\_expansion} $\gets$ true\;
                }
            }
        }
        
        \If{NOT \var{has\_expansion}}{
            \textbf{break}\; \Comment{\footnotesize No beams could be expanded}
        }
        
        sort \var{beams} by descending probability\;
        \var{beams} $\gets$ top \var{beam\_width} from \var{beams}\;
    }
    
    \var{next\_object} $\gets$ next object from highest probability beam in \var{beams}\;
    \var{sorted\_sequence} $\gets$ \var{sorted\_sequence} $\cup$ \{\var{next\_object}\}\;
    remove \var{next\_object} from \var{unsorted\_objects}\;
}

\KwRet{\var{sorted\_sequence}}\;
\end{algorithm}

\subsection{Placement Prediction}
\label{subsec:ppredition}

We consider two different paradigms to train the placement prediction policy. 
As such, the first consists of two distinct IL models, BC models.
The second approach relies on an RL-trained policy.

\subsubsection{Observation and Action Spaces}

An image of a box heightmap is used as the basis for observation of the considered models (as common in the literature~\cite{huang_planning_2023, zhao_learning_2023}).
A top-down projection of the next object to be packed is appended to the box heightmap, and then the complete image is padded to $224\times224$ pixels representing the observation $o$ (as in Fig.~\ref{fig:bcmse}).

The model target (action $a$) is a three-dimensional vector representing the object's $x$ and $y$ position and the planar rotation $\theta$. 
When predicting the pose, the vertical coordinate $z$ can be estimated from the box heightmap and the object model~\cite{huang_planning_2023, zhao_learning_2023}, thus reducing the action space.
The motivation behind constraining the rotation to be planar is twofold.
Firstly, it is challenging to reorient objects in a practical robotic scenario, especially while minding possible collisions~\cite{pantoja-benavides_comprehensive_2024, zhao_learning_2023}.
Secondly, by increasing the dimension of the action space, the learning problem would become increasingly complex, potentially without an increase in performance at a given task (observing results in Sect.~\ref{sec:res}).

In BC models, to represent the cyclic nature of the variable $\theta$ (especially under the limited amount of supervised data), the rotation is encoded into $\sin$ and $\cos$ components:
\begin{equation}
    \theta_{\sin} = \sin \left( \theta \right)
    \\
    \theta_{\cos} = \cos \left( \theta \right)
\end{equation}
In the IL experiments, $x$ and $y$ are normalized in $(-1,1)$. 
In the RL experiment, for simplicity in action sampling and without data amount constraints, all three output variables are normalized in $(-1,1)$.

\subsubsection{Imitation Learning}

\begin{figure}
    \centering
    \includegraphics[width=\linewidth]{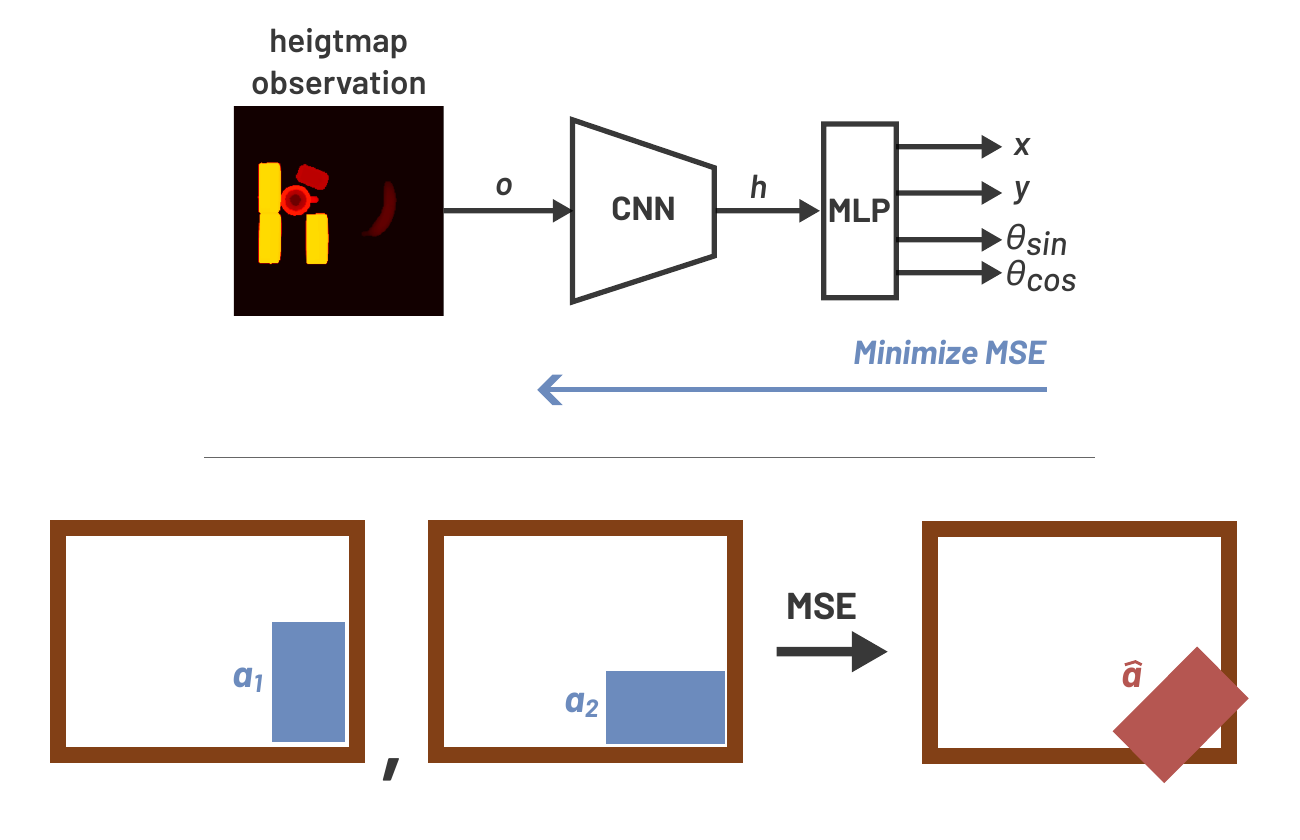}
    \caption{(Top) BC-MSE model that, given the heightmap observation of the box, predicts the placement pose, trained using MSE loss. (Bottom) Given the two dataset instances of valid placements ($a_1$ and $a_2$, $a \in [x,y,\theta]$) of the same object (in blue), the resulting MSE leads to bad placement prediction ($\hat{a}$, in red).}
    \label{fig:bcmse}
\end{figure}

If a \textit{naive} BC approach is considered, a model that maps the state of the box and the object to be placed to the target placement pose can be trained by minimizing the Mean Squared Error~(MSE) loss (we refer to this model as BC-MSE, Fig.~\ref{fig:bcmse}).
A CNN~\cite{mnih_human-level_2015} is used to extract the features from the observation image and encode them into a latent vector $z$. 

This approach fails due to the multimodality of human demonstrations and the inadequacy of MSE loss to capture the underlying distribution.
If Fig.~\ref{fig:bcmse} is observed, two placements represent two valid demonstrations of packing the same object.
However, the resulting pose optimizing MSE would be a bad placement, leading to the object tipping over.

\begin{figure}
    \centering
    \includegraphics[width=\linewidth]{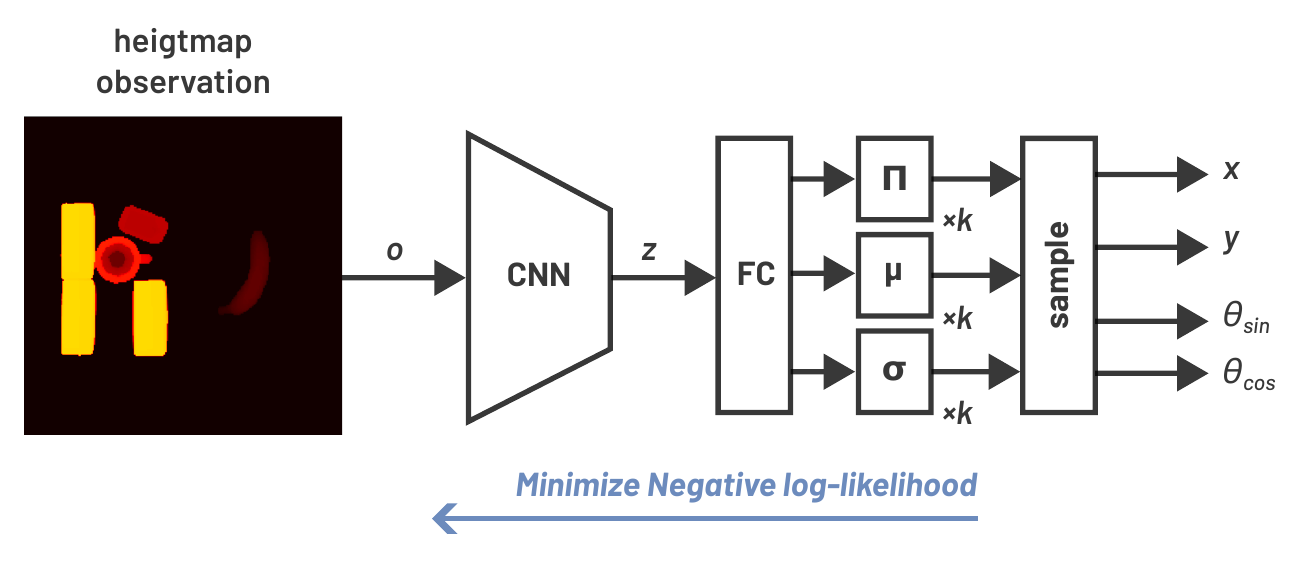}
    \caption{BC-MDN model that, given the observation of the box and the projection of the next object to be placed, outputs the mixture of Gaussians modeling the placement prediction trained on negative log-likelihood loss.}
    \label{fig:bcmdn}
\end{figure}

To address the \textit{multimodal} (referred to as multi-valued in the original paper) mappings, Bishop~\cite{bishop_mixture_1994} proposed Mixture Density Networks~(MDN).
Thus, multimodal distribution can be approximated as a mixture of Gaussian distributions:
\begin{equation}
    P(a|h) = \sum^K_{k=1} \Pi_k(h) \phi(a, \mu_k(h), \sigma_k(h))
\end{equation}
where $k$ is an index of $K$ total Gaussians, $\Pi_k(h)$ is the summation weight (s.t. $\Sigma^K_{k=1} \Pi_k(h) = 1$), and $\phi$ is the Gaussian distribution represented by its mean $\mu_k(h)$ and standard deviation $\sigma_k(h)$.
Where $\Pi_k, \mu_k, \sigma_k$ are parametrized by NN (for a given $h$) for each $k$.
\textit{Softmax} operator is used to ensure that $\Sigma^K_{k=1} \Pi_k(h) = 1$.
Again, CNN~\cite{mnih_human-level_2015} can be used as a feature extractor.
To train the model, the negative log-likelihood between the estimated distribution and target data is used as the loss function:
\begin{equation}
    L_{MDN}(a|h)=-\log[\sum^K_{k=1}\Pi_k(h)\phi(a, \mu_k(h), \sigma_k(h))]
\end{equation}
To sample the distributions, \textit{Gumbel softmax sampling} can be used, resulting in a complete BC-MDN prediction model (Fig.~\ref{fig:bcmdn}).

While MDN are a powerful tool in approximating multimodal distributions and similar models have been successfully used for learning from trajectories~\cite{mandlekar_what_2022}, they underperform at placement prediction in a packing task (from results in Sect.~\ref{sec:res}).
This is because the model is not able to provide precise estimations (similarly discussed in~\cite{chi_diffusion_2024}).
The accurate prediction is crucial when placements (represented as exact choices in continuous space) are considered.
Further discussion on this follows in Sect.~\ref{sec:disc}.

\subsubsection{Reinforcement Learning}

To address the limitations of the IL approach, the placement prediction policy can be trained by RL.
\textit{Soft Actor Critic} (SAC) is employed as the backbone RL algorithm~\cite{haarnoja_soft_2018} for placement prediction learning.
A solid algorithm backbone is integral to focus on task-specific hyperparameter settings~(such as sequence planning and reward function) while keeping the algorithm hyperparameters fixed.
Thus, SAC is chosen due to its robustness to hyperparameter settings, its ability to perform continuous predictions, and relatively fast convergence~(compared to other policy gradient methods).
Again, to avoid any potential bias due to different encoder schemes, the same feature extractor is used for this model as well~\cite{mnih_human-level_2015}.

\begin{figure*}[t]
    \centering
    \includegraphics[width=\textwidth]{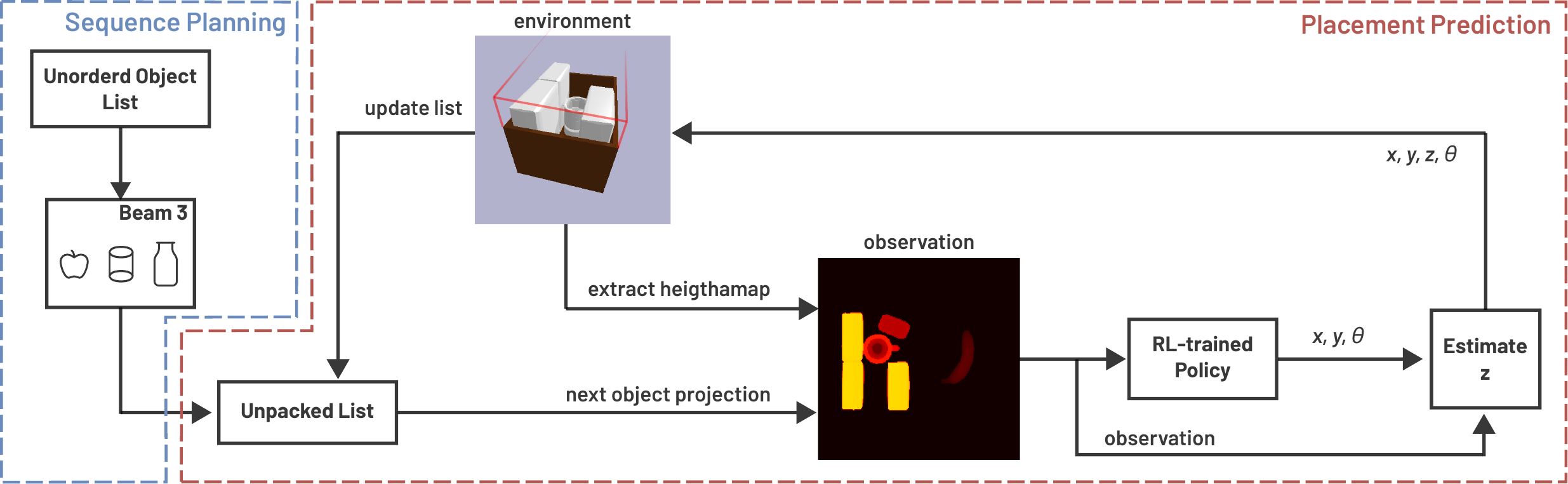}
    \caption{Complete block diagram of the HERB system in inference. Given an unordered list of objects to pack, the Beam-3 algorithm sorts the candidates (in blue). Continuously during packing (in red), the projection of the next object to pack is concatenated to the state of the box represented by a heightmap to construct the observation for the policy model. The policy predicts the $x,y,\theta$, which are then used to estimate the vertical position $z$ from the observation. The episode terminates upon the successful packing of all the objects, or if an object exceeds the sides of the box or the vertical constraint (here denoted by red lines in the 3D simulation of the box).}
    \label{fig:block_diagram}
\end{figure*}

We consider three reward functions.
The \textit{Simple} reward denotes a reward of 1 if the object is placed inside the box, and $-1$ otherwise.
For the second reward modality, \textit{Compactness} (sometimes referred to as \textit{utility}) of packing is considered.
While it can be defined differently across literature~\cite{pan_sdf-pack_2023, huang_planning_2023, zhao_learning_2023}, we adopt a definition akin to~\cite{huang_planning_2023} as it is the most general one and compatible with the BoxED dataset~\cite{santos_andrejfsantos4boxed_2025}:
\begin{equation}
    C = \frac{V_{cml}}{V_{min\_box}}
\end{equation}
where $V_{cml}$ is the cumulative volume of objects placed inside the box, and $V_{min\_box}$ is the volume of the minimal box encapsulating them.
This definition supports the dataset~\cite{santos_andrejfsantos4boxed_2025}, and thus human placements, since it is expected that the taller objects will overcome the height of the box.
Thus, the agent is rewarded $C$ if the object is placed inside the box, and -1 otherwise.
The third reward combines \textit{Compactness and Stability (CS)}, since a purely stability-driven reward yields sparse, inefficient packing. 
We define \textit{Stability (S)} as violating the rotation constraint (tipping beyond $10\degree$). 
Thus, the agent receives an additional reward \textit{S} of 1 if the constraint is respected, and 0 otherwise.
The CS reward is then composed as:
\begin{equation}
    CS =
    \begin{cases}
    \alpha C + (1 - \alpha) S & \text{if object\_inside} \\
    -1 & \text{else}
    \end{cases}
\end{equation}
where hyperparameter $0 \leq \alpha \leq 1$ is used to trade off compactness and stability.

A complete block diagram of the proposed approach is shown in Fig. \ref{fig:block_diagram}.

\section{EXPERIMENTAL SETUP}
\label{sec:exp}

\subsection{Dataset and Environment}

The environment for learning and evaluation is based on~\cite{santos_learning_2024}, however, while the original environment was implemented in \textit{Unity} physics engine, we decided to re-implement it in \textit{PyBullet}.
This is done to better exploit parallel simulations, facilitating faster RL training.
To enable the reproduction of human placement from BoxED~\cite{santos_andrejfsantos4boxed_2025}, appropriate transforms are applied. 
To further enhance the simulation speed, meshes from the dataset are simplified to be convex and watertight~\cite{zhao_learning_2023, stutz_learning_2020}.
The simulation is wrapped as a \textit{Gymnasium} environment.

When framed within the RL context, each step represents one object to be packed, and each episode represents one experimental session defined by the BoxED experiment~\cite{santos_learning_2024}.
The episode is terminated on successful placement of all the available objects or on an unsuccessful~(e.g., out of the box) placement of any object.
Since the height of the original box~(16.4~cm) is shorter than certain dataset objects, a vertical margin of 13 cm is added to the successful placement bounding box~(Fig.~\ref{fig:block_diagram}).

\subsection{Evaluation Procedure and Metrics}

We employ BoxED~\cite{santos_andrejfsantos4boxed_2025} trials and packing sequences as the test episodes for our experiments.
As such, the performance of different methods is evaluated over the same 263 experimental conditions (episodes) humans performed in~\cite{santos_andrejfsantos4boxed_2025}.
We evaluate the performance of different packing approaches and compare them with human packs based on the following objective task performance metrics:
\textit{Success rate}, which is defined as the percentage of successful test episodes, \textit{Number of packed objects}, which represents a distribution of packed objects over test episodes, and \textit{Latency}, the single-step inference time distribution.

We also report \textit{Compactness} and \textit{Stability} metrics across the test episodes.
While these metrics are qualitative, they provide significant insight into the performance of different algorithms and human behavior while packing.

\section{RESULTS}
\label{sec:res}

\subsection{Model Selection}

\subsubsection{IL}

To train the IL models, the BoxED dataset~\cite{santos_andrejfsantos4boxed_2025} is used.
The total number of object placements is 4644.
The placements that a planar rotation cannot approximate are excluded, leading to 4123 total data points.
These are divided into training (85\%) and validation (15\%) sets, and hyperparameter optimization is run separately for BC-MSE and BC-MDN models.
For BC-MSE, the size of the decoder network, the activation function of the decoder, the dropout rate, and the learning rate are optimized.
For BC-MDN, the same hyperparameters are optimized, along with the number of Gaussian components $K$.

\subsubsection{HERB}
\begin{figure}
    \centering
    \includegraphics[width=\linewidth]{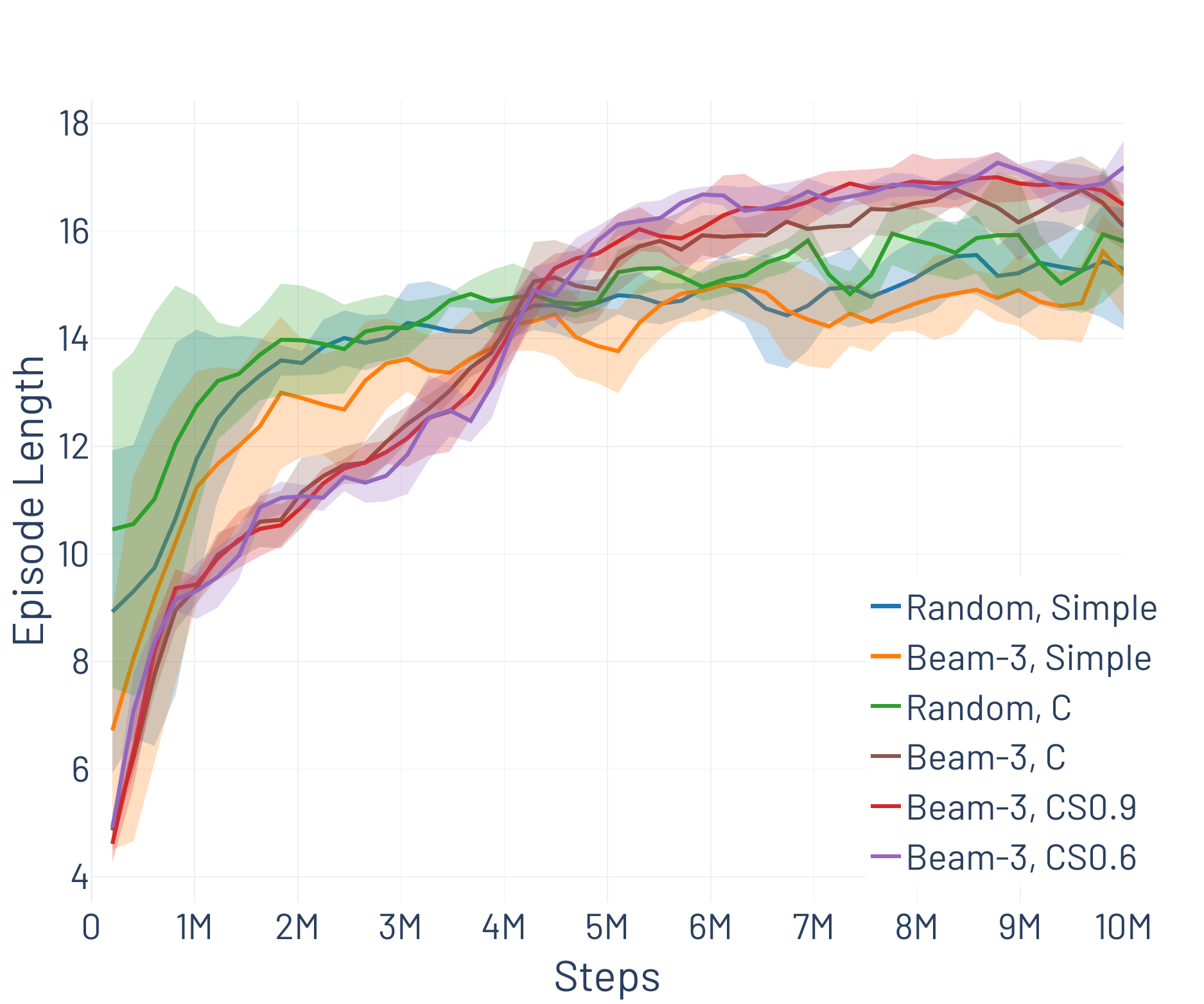}
    \caption{Episode Length (objects packed) mean and standard deviation over three seeds for each RL setting. Higher is better.}
    \label{fig:ablation}
\end{figure}

We study the impact of several design choices by running different flavors of HERB.
To facilitate the training of different variations, random episodes are initialized in an RL fashion.
These variations include the importance of selecting a random sequence of objects vs. Beam-3, and different reward functions introduced in Sect.~\ref{subsec:ppredition}.
Fig.~\ref{fig:ablation} reports the Episode Lengths~(equivalent to the number of objects packed) across different settings.
As the two best-performing parameter settings,  CS0.9 and CS0.6, display comparable performance in terms of RL metrics, they will be further analyzed in the following sections. 

All models were trained on an \textit{Ubuntu 22.04.5} with \textit{Intel Xeon Max 9460} CPU and \textit{NVIDIA A30 PCIe} GPU.
On this computer, training one RL seed for 10M steps takes $\sim$9 hours.
IL trainings are much shorter ($\sim$30 minutes), however, multiple runs for hyperparameter optimization would lead to a similar resulting time ($\sim$9 hours).

\subsection{Task Performance and Qualitative Metrics}

In this subsection, tests are performed on packing sequences from the BoxED~\cite{santos_andrejfsantos4boxed_2025} and compared to human performance.
Table \ref{tab:results} reports the task performance metrics across different methods and human packing.

As a baseline comparison, PackIt Heuristic~\cite{goyal_packit_2020} is used, which consists of sequence planning by the largest object, rotation alignment, and Bottom-Left-Back Fill (BLBF) position selection.
We consider two implementations of PackIt Heuristic, the SO(2), which limits the orientation to planar rotations, and the SO(3)~\cite{goyal_packit_2020}.

To evaluate the IL approaches, the sequence from the dataset is replayed to further remove negative bias due to possible limitations in sequence prediction.
Also, it is worth noting that there could be some positive bias in the performance of these models, as part of the BoxED dataset is used as the training set for these algorithms while also serving as the test set of this experiment.
However, even with these ``unfair advantages'' these models are suboptimal for the discussed task, and the omissions are deliberate to further the discussion.

To evaluate latency, the execution time of the placement prediction of the complete pose is considered.
Concerning the HERB algorithm and BC models, latency includes the Policy and Estimate z submodules (Fig.~\ref{fig:block_diagram}).
PackIt Heuristic SO(2) latency only includes the  BLBF module (since the objects are already rotation aligned).
Finally, PackIt Heuristic SO(3) latency includes the rotation alignment and BLBF modules.
The inference is run on a \textit{Ubuntu 20.04.6} machine with \textit{Intel i7-900} CPU and \textit{NVIDIA GeForce RTX 2060} GPU.

\begin{table}
\centering
\caption{Task performance metrics for different packing approaches (\textbf{BOLD} is the best, \underline{underline} second best).}
\label{tab:results}
\begin{tabular}{@{}lclrlrl@{}}
             & \multicolumn{2}{c}{}                      & \multicolumn{2}{c}{No. Objects}  & \multicolumn{2}{c}{Latency {[}ms{]}} \\
             & \multicolumn{2}{c}{Success {[}\%{]}} & mean           & (std)           & mean             & (std)             \\ \midrule
\textit{Target}    & \multicolumn{2}{c}{100}                   & 17.40          & (3.63)          & -                & -                 \\
Human Packing~\cite{santos_andrejfsantos4boxed_2025}       & \multicolumn{2}{c}{84.41}                 & 15.22          & (4.79)          & -                & -                 \\
PackIt SO(2)~\cite{goyal_packit_2020} & \multicolumn{2}{c}{82.50}                 & 16.33          & (4.46)          & 1.59             & (0.28)            \\
PackIt SO(3)~\cite{goyal_packit_2020} & \multicolumn{2}{c}{55.89}                 & 13.92          & (5.07)          & 2.51             & (0.52)            \\
BC-MSE & \multicolumn{2}{c}{7.22}                 & 6.69          & (5.16)          & 0.86             & (0.25)            \\
BC-MDN & \multicolumn{2}{c}{13.69}                 & 6.42          & (6.05)          & 0.85             & (0.31)            \\
HERB CS0.9      & \multicolumn{2}{c}{\textbf{87.83}}        & \textbf{16.76} & \textbf{(3.79)} & \underline{0.83}             & (\underline{0.26})   \\
HERB CS0.6      & \multicolumn{2}{c}{\underline{86.31}}                 & \underline{16.43}          & (\underline{4.26})          & \textbf{0.77}    & \textbf{(0.24)}           
\end{tabular}
\end{table}

\begin{figure}
    \centering
    \includegraphics[width=\columnwidth]{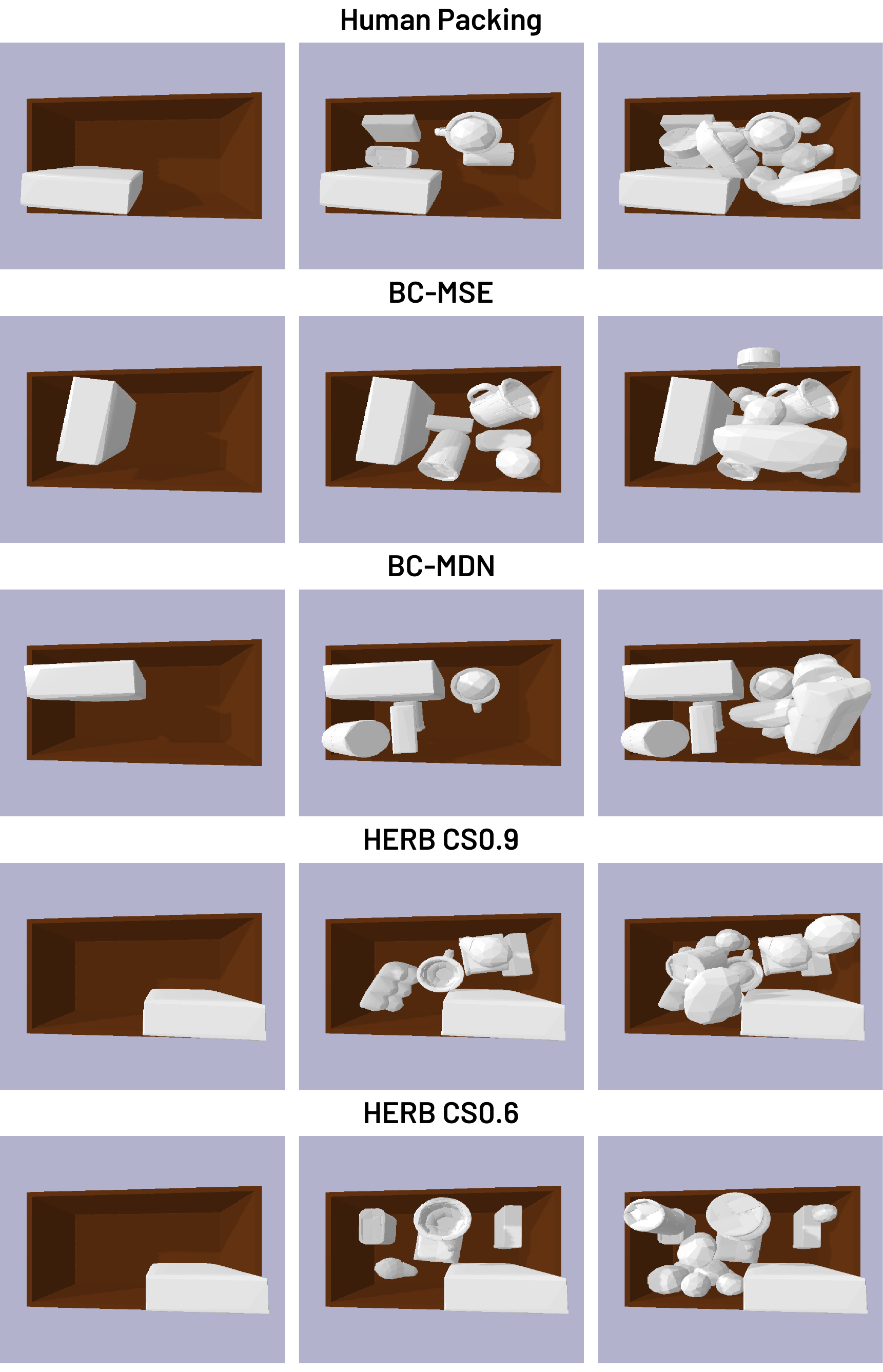}
    \caption{From left to right, the same pack from the BoxED dataset across human placements and data-driven models.}
    \label{fig:models}
\end{figure}

\begin{figure}[t]
    \centering
    \includegraphics[width=\linewidth]{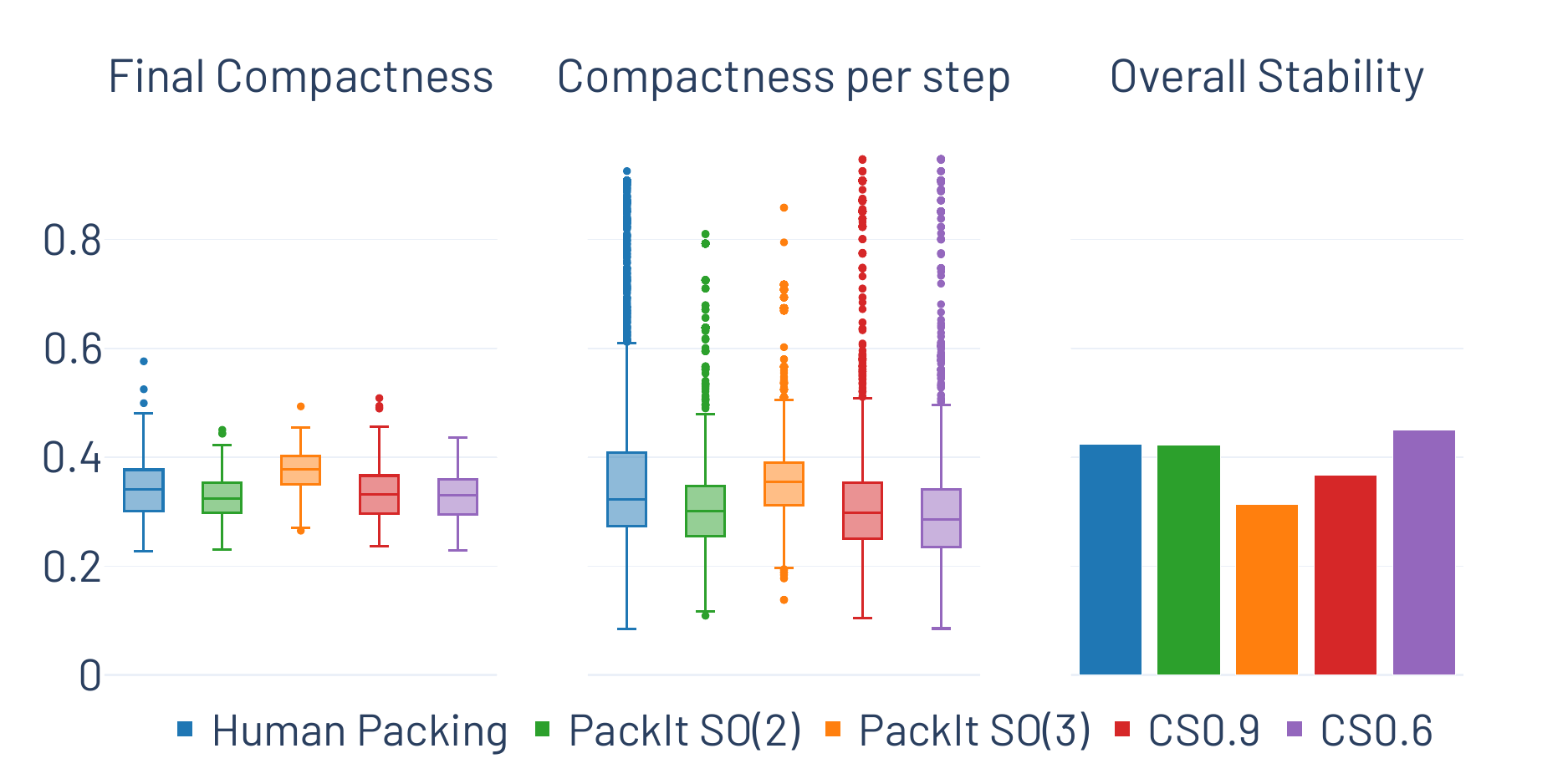}
    \caption{Final Compactness of the box (per successful episode), Compactness per step, and Overall Stability (whole dataset) for different approaches on the BoxED sequences.}
    \label{fig:cs}
\end{figure}

Fig.~\ref{fig:models} shows human and different models packing across one episode for qualitative comparison.
Additionally, Fig.~\ref{fig:cs} shows the compactness and stability distributions.
To estimate the stability of human placements and PackIt Heuristic SO(3), the same threshold of $10\degree$ is considered between the intended pose of the object and the predicted one.

\subsection{Robotic System}
\label{sec:robotic}

To validate the feasibility of the proposed method, a packing robotic system is set up (Fig. \ref{fig:setup}).

\begin{figure}
    \centering
    \includegraphics[width=\columnwidth]{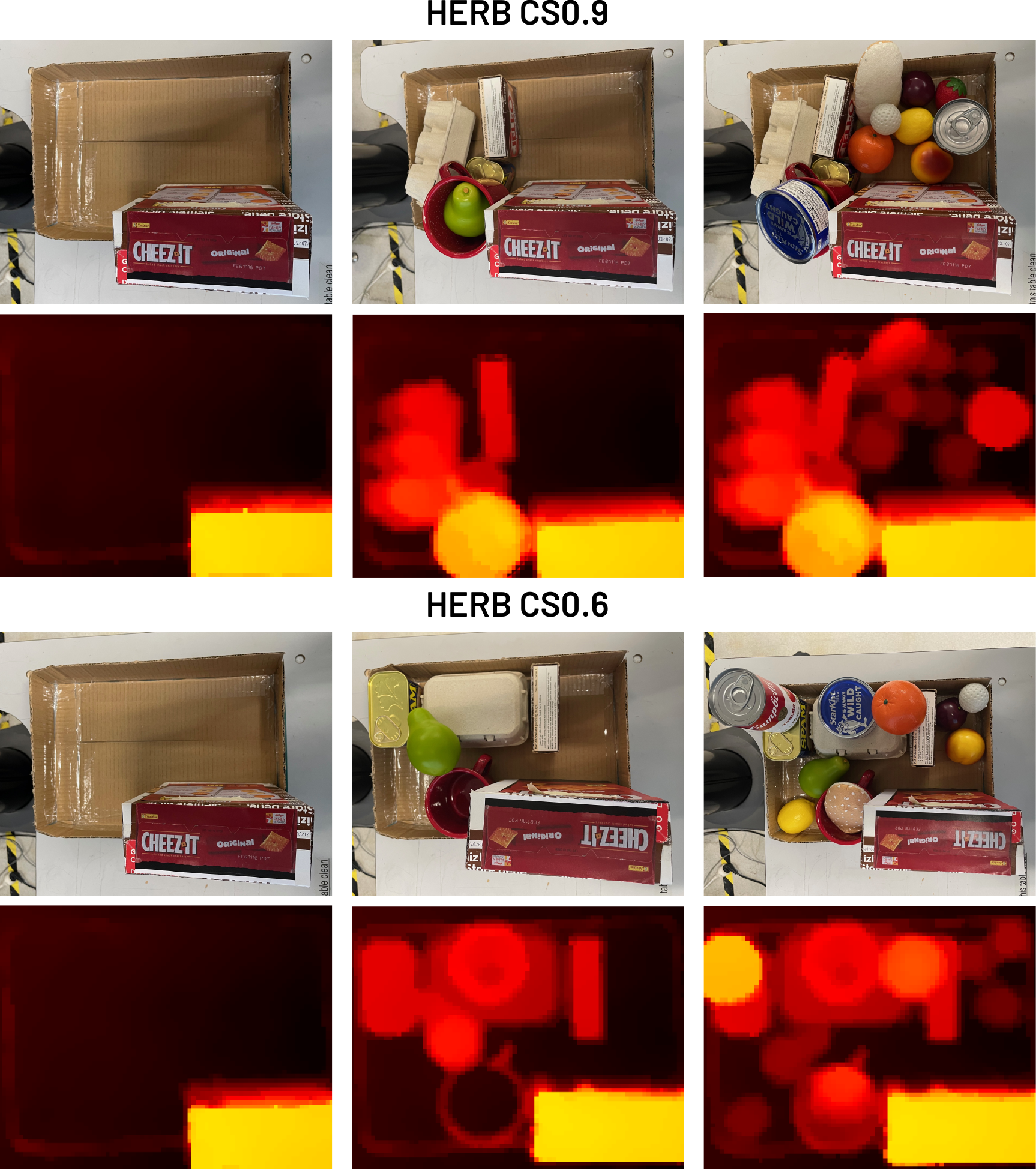}
    \caption{Comparison of placement prediction across CS0.9 and CS0.6 parameter settings on a BoxED pack. The top rows are the current state of the box, the bottom rows are the actual heightmaps that are used to construct the observation for the placement prediction model.}
    \label{fig:packs}
\end{figure}

To obtain the box heightmap, \textit{Intel \textregistered\ RealSense \texttrademark\ L515} LiDAR camera is used.
The image obtained from the sensors is cropped, padded, and threshold-filtered to mitigate noise around the bottom of the box. 
The two best-performing models (HERB CS0.9 and HERB CS0.6) are qualitatively assessed to determine their robustness to real-world input and predicted poses.
A demonstration of this assessment is performed by perturbing the predicted poses under real sensory input~(Fig.~\ref{fig:packs}).
Following this assessment, we chose CS0.6 as the preferred parameter setting for operation.
We further discuss our findings in Sect.~\ref{sec:disc}.

A Baxter\textregistered\ bi-manual robot equipped with two electric parallel grippers is selected as our robotic platform.
The range of one gripper is adjusted to grasp smaller objects, while the other is adjusted for larger objects from the dataset.
We used \textit{ROS Noetic} to facilitate picking, placing, and interfacing with HERB.
A video demonstrating the HERB system performance is included as supplementary material.

\section{DISCUSSION}
\label{sec:disc}

\subsection{Experimental Results}

From the ablation studies (Fig.~\ref{fig:ablation}), it can be concluded that both the C and CS reward modalities and human-like sequence planning benefit the RL approach.
It is worth noting that models trained with random sequences exhibit competitive performance, particularly early in the training.
This is expected, as they quickly learn to pack more objects~(often smaller ones) and adapt to suitable placement strategies.
In contrast, Beam-3, which is trained on human data, tends to prioritize packing larger objects first.
While this approach can be more prone to failure due to suboptimal placements, Beam-3 gradually improves over time.
As training progresses, it catches up and eventually surpasses other methods by learning to efficiently pack an increasing number of objects.
Furthermore, while the different settings of the parameter $\alpha$ seem to lead to similar RL performance (with marginal improvements with CS0.6 and CS0.9), having a stability component is important for downstream robotic tasks, as it reduces the detriment of collisions between objects while placing.
Qualitatively (Fig.~\ref{fig:models} and Fig.~\ref{fig:packs}), it also incentivizes a more human-like placing by keeping objects in their canonical orientations (i.e., not flipping bottles or egg cartons).

If Table \ref{tab:results} is considered, the benefit of the proposed approach in terms of decision latency can be observed.
When compared to PackIt Heuristic~\cite{goyal_packit_2020}, the shorter inference time of the proposed method is notable.
Furthermore, by simplifying the RL pipeline, a significant speed-up is also obtained in training time.
For example, related work such as \cite{huang_planning_2023} and \cite{zhao_learning_2023} report $\sim$16 GPU hours and $\sim$48 hours, respectively (compared to proposed $\sim$9 hours).
Still, the main bottleneck remains the simulation time, as the model inference and update are comparably much faster than simulating contacts in the box and ray casting to obtain the heightmap.

Conversely, the IL-based approaches (BC-MSE and BC-MDN) lead to poor performance (Table~\ref{tab:results}).
A combination of factors likely leads to this.
As mentioned, BC learning using MSE is especially ill-posed due to the multimodal nature of the problem.
The BC-MDN model, while exhibiting deficient performance, manages to capture some qualities of human packing.
For example, in the demonstrated pack (Fig.~\ref{fig:models}), the BC-MDN places the fruit inside the mug.
This is a space-optimizing strategy humans exploit as well (Human Packing in Fig.~\ref{fig:models}).
Still, the lack of precision of the probabilistic approach leads to many failures, as the placement precision is critical in this task.
Qualitatively, both models eventually ``average out'' the placement targets.
The BC-MSE does this by tending to place objects in the center of the box.
On the other hand, BC-MDN does this by replicating the common factor across all the trials, that is, placing objects somewhere in the empty space.
However, since it does not align the objects accurately, it still leads to poor performance and objects spilling out.

This lack of precision exhibited by IL approaches also highlights the difficulties in learning from precise human choices in continuous space (as opposed to more common trajectory learning).
As such, imprecision leads to compounding errors, detrimental to the performance of the approach (as seen in Table~\ref{tab:results} and Fig.~\ref{fig:models}).
Recent models such as~\cite{chi_diffusion_2024} could help alleviate this, as they tend to model multimodal distributions with higher precision.
However, more complex models could be arduous to train, requiring more data, while the combination of human-like placement prediction and RL-based placement policy integrates seamlessly, slightly outperforming even the human packing.
As such, the proposed method is an efficient and effective solution for the considered benchmark, and a more elaborate modeling of human placement will be evaluated in future work, which will aim to overcome some limitations imposed by the dataset (such as known objects and box size).

When examining human performance in terms of Success Rate and Number of packed objects~(Table~\ref{tab:results}), it can be noted that in the original experiment~\cite{santos_learning_2024} participants do not always adhere to the assumptions made by the proposed RL environment.
When examining the dataset~\cite{santos_andrejfsantos4boxed_2025}, some participants would place some objects leaning over the edge of the box or overcome the assumed vertical limit.
This leads to somewhat worse performance than expected when the participants' performance is considered.

Although the PackIt Heuristic SO(2) obtains comparable performance in terms of objects packed, it is somewhat limited by the discretization of the environment, which leads to more unreliable packing, resulting in 5\% drop in Success Rate.
PackIt Heuristic SO(3)'s poor performance highlights the shortcomings in transferring heuristic assumptions to realistic scenarios.
By ordering objects by volume, larger objects that are not cuboidal (egg carton, bread, bleach bottle, mustard bottle) are placed at the start of the packing somewhat non-canonically, i.e., by aligning the longest dimension with the longest side of the box.
In doing so, the resulting packs rely on an unstable base, and even though the initial placements tend to be compact~(Fig.~\ref{fig:cs}), the later objects are unstable and fall out of the box.
This is also reflected when Stability results are considered~(Fig.~\ref{fig:cs}).
Furthermore, qualitatively, this could be deemed as poor packing since these objects could be fragile or prone to spilling.
This further highlights the potential of obtaining useful information from humans implicitly, as both task performance and qualitative improvements can be achieved.

If Final Compactness and Compactness per step in Fig.~\ref{fig:cs} are examined, it can be concluded that different methods (and humans) have exhibited high variance while placing, converging to similar values as the box gets packed.
While compactness can serve as a proxy for the packing utility \cite{zhao_learning_2023}, optimizing other factors (such as stability or the quality of the sequence) can be improved without compromising it.
The stability constraint thresholded by $10\degree$ could be slightly conservative, leading to low stability metrics across the board.
Furthermore, spherical objects~(of which there are plenty in a household items dataset such as BoxED) tend to roll under collisions.
Finally, as humans do not seem to mind these rotations while placing objects (Fig.~\ref{fig:cs}), this input could be used in designing a future system that could care about the stability of certain objects more than others.

\subsection{Robotic System}

Fig. \ref{fig:packs} shows that HERB is capable of working with data from a real depth camera in a zero-shot manner, and without any fine-tuning.
The model is robust to imperfect placements and real-world physics.
As expected, due to the sim-to-real gap, the execution of the model in the real world (Fig.~\ref{fig:packs}) diverges from the same sequence run in simulation (Fig.~\ref{fig:models}). 
However, the prediction model remains within its learned distribution, successfully estimating poses that lead to a complete and stable packing arrangement.

Although prioritizing compactness can, in theory, maximize the number of objects packed, our robotic experiments (Sect.~\ref{sec:robotic}) indicate that a balanced trade-off between \emph{compactness} and \emph{stability} yields better results~($\alpha = 0.6$, i.e. CS0.6).
For example, in Fig. \ref{fig:packs}, the egg carton flips in the CS0.9 setting, which could be due to the RL agent exploiting other objects to enhance compactness, something that results in a significant hindrance to the quality of the packing.

The robotic system effectively executes sequences provided by BoxED. Human-guided sequences tend to prioritize placing stable objects first, creating a more reliable foundation for later placement of unstable or spherical objects, thereby improving overall packing stability.
Importantly, by learning the continuous pose through RL, the model implicitly acquired experience with challenging or potentially unstable configurations, adding a layer of reliability.
Bin picking and packing in cluttered environments are very challenging problems, particularly from the dexterity point of view.
By using constraints such as pre-determined grasps and top-down placements, these challenges can be mitigated but not completely alleviated.
While addressing collisions and trajectory planning is out of the scope of this work, placement planning in continuous space is an important step towards more dexterous and versatile robotic agents.

\section{CONCLUSIONS}
\label{sec:conc}


We propose HERB, a framework that combines RL with human-like sequence prediction, and demonstrate its effectiveness for reliable 3D irregular object packing. 
The human-like sequence prediction not only reduces the complexity and training time of the RL approach but also incorporates latent information about object properties (such as fragility, deformability, and human preferences) that are often overlooked. 
Our results outperform state-of-the-art benchmarks, and experiments on a real robotic platform validate the feasibility of the method in practical scenarios, highlighting the importance of incorporating the notion of stability into packing planning.

Looking forward, we will focus on advancing continuous bin packing and developing more dexterous solutions, potentially including grasp planning. We also plan to explore Vision-Language Foundation Models to reason about relationships between previously unseen objects and generate adaptive packing accordingly.

By framing irregular object packing in a continuous robotic setting, HERB opens the door to more robust, efficient, and human-aware manipulation systems, bridging the gap between simulation and real-world deployment.








\bibliographystyle{IEEEtran}
\bibliography{references}

\end{document}